\documentclass{article}

\usepackage[preprint]{neurips_2026}

\usepackage[utf8]{inputenc} 
\usepackage[T1]{fontenc}    
\usepackage{hyperref}       
\usepackage{url}            
\usepackage{booktabs}       
\usepackage{amsfonts}       
\usepackage{nicefrac}       
\usepackage{microtype}      
\usepackage[table]{xcolor}  
\usepackage{graphicx}         
\usepackage{amsmath}  
\usepackage{multirow}  
\usepackage{caption}
\usepackage{wrapfig}
\usepackage{algorithm}
\usepackage{algpseudocode}
\usepackage{booktabs}
\usepackage{amssymb}  
\title{Large Language Model as Token Compressor and Decompressor}

\author{Wenbing Li\\
\and
Yiran Wang\\
\and
Zikai Song\\
\and
Jielei Zhang\\
\and
Tianhao Zhao\\
\and
Junkai Lin\\
\and
Huipeng Guo\\
\and
Wei Yang$^{*}$\\
}

\begin{document}

\maketitle

\vspace{-1em}
\begin{abstract}

In this paper, we study whether an off-the-shelf LLM can be adapted into a discrete, variable-length token compressor and decompressor for long-context processing. To this end, we design a self-expressive autoencoding framework that fine-tunes a pretrained LLM with lightweight LoRA adapters to map long texts into compact sequences of learned latent codes, termed Z-tokens, and to decode them back into natural language or task outputs. The resulting representation is content-adaptive: less predictable or information-dense segments can receive more Z-tokens, while redundant regions can be represented more compactly through a budget-aware length regularizer.
Our method is evaluated on long-context datasets such as Wikipedia, CNN/DailyMail, HotpotQA, and QuALITY, showing that it preserves reconstruction quality and downstream performance while reducing effective context length, generation-stage memory usage, and end-to-end latency. This simple design supports both direct decoding from compressed contexts and autoregressive generation in the Z-token space, providing a practical interface for efficient long-context inference.
\end{abstract}

\vspace{-1em}
\section{Introduction}
\vspace{-1em}

Large language models (LLMs) have achieved remarkable success in various fields~\cite{llmfewshotlearners, llama}. However, their ability to handle long contexts remains fundamentally limited by the quadratic complexity of the attention mechanism~\cite{attention}. As sequence lengths grow to tens or even hundreds of thousands of tokens, the computational and memory overhead of the attention mechanism makes large-scale inference, long document understanding, and multi-hop retrieval prohibitively expensive. 
To mitigate this, significant research has focused on token compression, which aims to distill lengthy sequences into shorter, information-dense representations~\cite{incontextae, distillingcontext, autocompressor, gisttokens}.
Existing approaches can be broadly categorized into two categories. Heuristic token merging methods, such as similarity-based merging~\cite{tokenmerging, aim}, attention-based pruning~\cite{tokenprune}, and saliency-guided aggregation~\cite{compressingcontextenhanceinference}, attempt to remove or merge redundant tokens based on local similarity or attention importance scores. While these techniques are effective in shortening sequence lengths, 
tokens that appear redundant due to similarity may carry discourse or pragmatic meaning that is difficult to recover. The second category includes learning-based or autoencoding methods such as AutoCompressor~\cite{autocompressor}, ICAE~\cite{incontextae}, or Gist Tokens~\cite{gisttokens}. AutoCompressor and ICAE use LLMs to generate fixed-length compressed context vectors that can be re-expanded for downstream tasks by training auxiliary networks or soft embedding modules. Gist Tokens force the model to compress cues into fixed-length lexical units through an attention masking mechanism. 
These methods often treat compression as an external, auxiliary task, decoupled from the LLM's primary generative reasoning process. 
%
Moreover, their reliance on fixed compression ratios prevents adaptation to varying information density and typically forgoes faithful reconstruction of the original input, limiting their applicability in scenarios that demand verifiability or fine-grained access to the source content.
In this work, we depart from prevailing heuristic and learned-embedding approaches by introducing a novel insight: leveraging the semantic knowledge of the LLM to serve as both a token compressor and decompressor. Instead of representing the compressed context as a continuous vector or a fixed number of memory slots, we expand the LLM vocabulary using the learned Z-token codebook and generate compressed tokens autoregressively. 
Our method is predicated on the fundamental insight that LLMs, trained on vast corpora, are not merely statistical models but powerful engines for semantic abstraction and conceptual distillation~\cite{llmemergent, foundationmodels, llmiscompression}. We observe that natural language is inherently compressive; it exhibits a strong tendency to encode complex ideas into compact lexical units. Idioms such as ``double-edged sword'' encapsulate intricate trade-offs, and metaphors like ``tip of the iceberg'' summarize extensive underlying complexities. This phenomenon implies that LLMs themselves are capable of efficient compression.
Methodologically, we propose an autoencoding training paradigm, where an input sequence $X$ is mapped into a compressed sequence of discrete Z-tokens, $Z$, by a LLM compressor, and subsequently reconstructed or continued by a LLM decompressor. The training objective can be viewed as a form of bilingual translation, where the model learns to encode natural language into its own abstract internal language and decode it back. We make this process explicit with autoregressive discrete generation over an extended codebook, plus a budget-aware length regularizer that controls the realized compression ratio. The compressed latent $\{ Z \}$ serves as a semantic scaffold, forcing the model to perform core reasoning on compact symbolic representations before projecting them back to surface tokens.
Our design is more flexible, capable of dynamically lengthening compressed text based on semantic density, and seamlessly integrating with sliding windows to support longer contexts. We further analyze the contextual regularity of the learned Z-token codebook and show that repeated Z-tokens tend to occur in semantically related contexts, while still remaining context-dependent rather than fixed lexical units.
We conducted extensive experiments on six large-scale datasets—Wikipedia, CNN/Daily Mail, HotpotQA, NarrativeQA, QASPER, and QuALITY—covering text reconstruction, continuation, summarization, and long text question answering, demonstrating the effectiveness of our method.
\vspace{-1em}
\section{Related Works}
\vspace{-1em}
\label{sec:relatedwork}
\begin{figure}[!t]
    \centering
    \includegraphics[width=\textwidth]{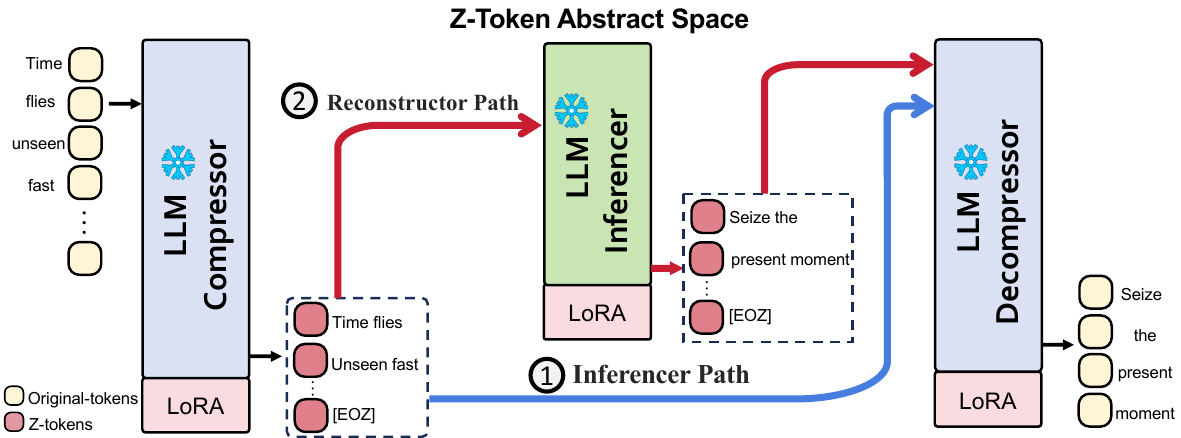}
    \caption{Our framework consists of three components: a compressor, an inference module, and a decompressor. It supports two usage paradigms. In the first, we directly feed the compressed representation to the decompressor to perform the downstream task. In the second, we first run the inference module in the Z-token space and then invoke the decompressor to recover task outputs at the token level.}
    \label{fig:system_overview_1}
    \vspace{-2em}
\end{figure}

\noindent \textbf{Context Compression}
refers to any mechanism that explicitly aims to reduce the number of tokens participating in subsequent computation.
Some approaches follow \textit{heuristic token reduction} strategies, removing or merging less informative tokens based on saliency or similarity. Token Merging~\cite{tokenmerging, liu2025catanetefficientcontentawaretoken, aim} aggregates similar tokens , while Token Pruning~\cite{tokenprune, aim, dynamicvit, compact, hierarchicalcontextmergingbetter} eliminates low-saliency tokens during inference. 
Beyond rule-based methods, \textit{learning-based semantic compression} encodes contexts into compact representations that retain high-level meaning. ICAE~\cite{incontextae} and AutoCompress~\cite{autocompressor} represent long sequences as latent vectors for efficient reconstruction or reasoning; other recent works similarly investigate learned context compression~\cite{distillingcontext, activationbeacon}.
A related branch, \textit{prompt compression}, targets input instructions rather than the full context. Methods such as 500xCompressor~\cite{500xcompressor}, Gist Tokens~\cite{gisttokens} and LongLLMLingua~\cite{longllmlingua} summarize long prompts into shorter, reusable representations.
Our method is conceptually aligned with \textit{learning-based semantic compression} but differs in execution: rather than encoding context externally, we integrate compression into the LLM’s reasoning process. This yields discrete internal symbols that form a self-expressive “semantic language,” achieving adaptive, interpretable, and semantically faithful compression.

\noindent \textbf{Architectural and Implicit Sparsity Methods} 
represent an alternative line of research that, instead of explicitly reducing the number of tokens involved in computation, improve efficiency through architectural modifications or implicitly introduced sparsity in the computation process. 
Compressive Transformer~\cite{compressive} introduces hierarchical memory buffers summarizing distant activations; Reformer~\cite{reformer} leverages locality-sensitive hashing to approximate attention; Routing Transformer~\cite{routing_transformer} sparsifies attention through content-based routing.
Other models such as Longformer, BigBird, and Window Attention~\cite{sparseattn, longformer, bigbird, deepseeknsa, maskllm, switchtransformer, deepseekmoe} also achieve scalability via structured sparsity.

\noindent \textbf{Discrete Semantic Representation}
provides a principled foundation for compact and interpretable encoding of high-level semantics. VQ-VAE~\cite{vqvae} introduced vector-quantized bottlenecks that map continuous signals into a learned codebook, later extended to sequence models~\cite{kaiser2018discrete, kaiser2018fast, liu2024discretesemantictokenizationdeep, zhao2018unsuperviseddiscretesentencerepresentation,song19} and cross-modal learning frameworks~\cite{liu2021crossmodaldiscreterepresentationlearning}. These approaches show that discrete latent variables can serve as symbolic carriers of information. Subsequent works explored training stability and optimization for discrete variables~\cite{jin2020discrete}.
While prior works used discrete bottlenecks as auxiliary modules, we internalize this semantic layer within the LLM’s reasoning flow, enabling discrete tokens to serve as learned, interpretable units for compressing and reasoning over long contexts.

\vspace{-1em}
\section{Methods}
\vspace{-1em}
Our goal is to adapt a pretrained LLM into a compressor and decompressor. The compression and decompression work as follows: given an input sequence $\mathbf{X} = \{ \mathbf{x}_1,  \mathbf{x}_2, \dots, \mathbf{x}_N\}$, the compressor generates a variable-length sequence of tokens $\mathbf{Z}  = \{\mathbf{z}_1, \mathbf{z}_2, \dots, \mathbf{z}_{K}\}$, where $K \ll	N$.
%
We consider two usage modes:
1). \textbf{Direct decompression for inference}: the decompressor directly consumes Z-tokens and generates downstream task outputs, so the compressor acts as a learned prompt compressor;
2). \textbf{Z-space inference followed by reconstruction}: the decompressor reconstructs surface text from Z-tokens, while a separate inferencer operates in the Z-token space before final decoding.
Since the compressor maps an $N$-token input to a much shorter Z-token sequence of length $K \ll N$, subsequent attention-based computation scales as $\mathcal{O}(K^2)$ instead of $\mathcal{O}(N^2)$, yielding a theoretical compute and memory saving of approximately $(N/K)^2$ and enabling faster inference, lower memory usage, and effectively longer contexts.

\vspace{-1em}
\subsection{LLM Compressor}
Given an input token sequence $\mathbf{X}$, the compressor generates a variable-length Z-token sequence $\mathbf{Z}$. To preserve the prior of the pretrained LLM, we extend the original vocabulary to $[\mathcal{V}_{\text{base}}, \mathcal{V}_z]$, where $\mathcal{V}_{\text{base}}$ is the original vocabulary and $\mathcal{V}_z$ is the extended vocabulary for Z-tokens. Z-tokens are generated only from $\mathcal{V}_z$. Unlike previous work (e.g., ICAE or AutoCompressor) that directly learns fixed-length soft embeddings, our compressor generates $\mathbf{Z}$ autoregressively:
\begin{equation}
    p_\phi(\mathbf{Z}\mid\mathbf{X}) = \prod_{t=1}^{K} p_\phi(z_t \mid \mathbf{X}, z_1, \dots, z_{t-1}).
\end{equation}

The compressed length $K$ is not fixed a priori but determined dynamically by the compressor itself. Concretely, we augment $\mathcal{V}_z$ with a special \texttt{[EOS-Z]} token; generation stops once the compressor emits \texttt{[EOS-Z]}, so $K$ emerges from the model output rather than being imposed as a hyperparameter. To avoid degenerate solutions in which the model either collapses to a single Z-token or reproduces the input length, we regularize the realized compression ratio against a target ratio $r$:
\noindent
\begin{minipage}[t]{0.48\linewidth}
    \vspace{-1em}
\begin{equation}
    \mathcal{L}_{\text{len}}=
    \left(\frac{K}{N}-\frac{1}{r}\right)^2
\end{equation}
\end{minipage}\hfill
\begin{minipage}[t]{0.48\linewidth}
    \vspace{-1em}
\begin{equation}
    z_t=\mathbf{GumbelSoftmax}(\mathbf{h}_t), \quad z_t\in\mathcal{V}_z
\end{equation}
\end{minipage}

where $N$ is the input length. This soft constraint allows informative segments to consume more Z-tokens while penalizing global budget violations, enabling content-adaptive allocation without imposing rigid per-segment quotas. This design can also be interpreted from a constrained information allocation perspective. Suppose $\mathbf{X}$ is partitioned into local segments $\{\mathbf{X}^{(i)}\}_{i=1}^{M}$, and the compressor allocates $K_i$ latent tokens to segment $\mathbf{X}^{(i)}$ under a global budget $\sum_i K_i \le N/r$. If the segment-wise information density is defined as
\[
H_i \triangleq H(\mathbf{X}^{(i)}\mid \mathbf{X}^{(<i)}),
\]
and the distortion for segment $\mathbf{X}^{(i)}$ takes the form $D_i(K_i)=H_i e^{-\beta K_i}$ with $\beta>0$, then the optimal allocation satisfies
\[
K_i^\star=\frac{1}{\beta}\bigl(\log(\beta H_i)-\log\lambda\bigr),
\]
for some $\lambda>0$, and is therefore monotonically increasing in $H_i$. In other words, under a fixed compression budget, semantically denser or less predictable segments should receive more Z-tokens, while more redundant segments should receive fewer. From this perspective, the \texttt{[EOS-Z]} mechanism together with $\mathcal{L}_{\text{len}}$ implements a soft rate--distortion trade-off. A complete derivation and discussion are provided in Appendix~\ref{thero}. 
Therefore, the model can dynamically determine the lexical units needed to represent a given context, achieving variable-length compression in a way that remains compatible with the native generation process of LLMs. To ensure end-to-end training while maintaining discreteness, we use a Gumbel-Softmax estimator to sample Z-tokens only from $\mathcal{V}_z$. This produces a soft embedding $e_{\text{soft}}$ during training. We then apply a straight-through estimator to combine differentiable gradients with discrete decoding:

\begin{minipage}[t]{0.48\linewidth}
    \vspace{-1em}
\begin{equation}
    e = e_{\text{hard}} + [e_{\text{soft}} - \mathbf{sg}(e_{\text{soft}})]
\end{equation}
\end{minipage}\hfill
\begin{minipage}[t]{0.48\linewidth}
    \vspace{-1em}
\begin{equation}
    \mathcal{L}_{\text{overlap}} = 1 - \cos(\mathbf{Z}_i^{\text{overlap}}, \mathbf{Z}_{i+1}^{\text{overlap}})
\end{equation}
\end{minipage}
where $e_{\text{hard}}$ denotes the embedding obtained by argmax. This hybrid estimator maintains stable gradient flow without sacrificing the discreteness of the learned symbols.

\noindent \textbf{Sliding window compression.} For excessively long sequences, our method naturally adapts to sliding-window compression: the input sequence is segmented into overlapping windows of length $W$ and stride $S$ (typically $W=1024$ and $S=256$), each independently compressed into $\mathbf{Z}_i$. For adjacent segments, we enforce a consistency constraint on overlapping regions to ensure semantic coherence. We can then concatenate tokens of compressed segments $\{ \mathbf{Z}_i \}$ for long-context reconstruction or inference.

\vspace{-1em}
\subsection{Direct decompression for inference}
\label{sec:decompressor}
The LLM decompressor serves as the core component for translating the compressed Z-tokens back to meaningful semantic outputs.
Our decompressor supports two distinct usage paradigms: the inferencer and the reconstructor.
For both paradigms, the decompression process is modeled as conditional autoregressive generation, with shared training objectives and constraints that ensure stability and semantic fidelity.
Formally, the decompression process follows the distribution:
\vspace{-0.5em}
\begin{equation}
p_\theta(\mathbf{Y}|\mathbf{Z}) = \prod_{t=1}^{T} p_\theta(y_t \mid y_1 \dots y_{t-1}, \mathbf{Z}),
\end{equation}
Depending on the usage paradigm, the sequence $\mathbf{Y}$ varies. For the reconstructor task, $\mathbf{Y}$ is the input sequence to the compressor $\mathbf{X}$. While for the inferencer task, such as Question-Answering (QA), $\mathbf{Y}$ is the response to $\mathbf{X}$. $\mathbf{Z}$ represents the Z-tokens generated by the compressor.
To ensure a linguistic foundation, the decompressor is trained under lexical constraints: in each decoding step, the output logit vector is restricted to within the pre-trained model's original vocabulary $\mathcal{V}_{\text{base}}$:
\begin{equation}
\tilde{p}_\theta(y_t|y_1 \dots y_{t-1}, \mathbf{Z}) = \textbf{Softmax}\left( W_{\text{base}} \mathbf{h}_t \right),
\end{equation}
where $\mathbf{h}_t$ represents the decoder's hidden state at step $t$, and $W_{\text{base}}$ is the embedding matrix of the original vocabulary set. This constraint forces the model to use human-understandable vocabulary to express internal abstract representations, effectively translating the private latent ``language'' back into natural language. Empirical evidence shows this vocabulary constraint improves training stability and ensures semantic fidelity across tasks. For training, the general form of the loss is a cross-entropy term that maximizes the likelihood of generating correct sequence given the Z-tokens:
\begin{equation}
\mathcal{L}_{\text{tr}} = - \sum_{t=1}^{T} \log p_\theta(y_t \mid y_1 \dots y_{t-1}, \mathbf{Z}),
\end{equation}
Training entirely under teacher forcing introduces exposure bias, where the model learns to rely on ground-truth labels that are unavailable at inference time. To mitigate this mismatch, we adopt a \emph{scheduled sampling} strategy: at each decoding step, the lexical input is taken from the ground-truth token with probability $1-p$, or from the model's own previous prediction with probability $p$. The probability $p$ increases linearly during training, allowing the model to gradually transition from teacher-forced decoding to free decoding. This improves robustness to its own prediction errors and leads to better generalization at inference time.

In addition to the token-level reconstruction objective, we introduce two regularization terms to stabilize discrete code learning and promote effective use of the Z-token vocabulary. The first term, $\mathcal{L}_{\text{com}}$, regularizes the consistency between the soft embedding selected by Gumbel-Softmax and its corresponding hard embedding:
\begin{equation}
\label{eq:lcom}
\mathcal{L}_{\text{com}} =
\mathbb{E}_{t}\!\left[\left\|\mathbf{sg}(\mathbf{e}_{t}^{\text{soft}})-\mathbf{e}_{t}^{\text{hard}}\right\|_2^2\right]
+\eta\,
\mathbb{E}_{t}\!\left[\left\|\mathbf{e}_{t}^{\text{soft}}-\mathbf{sg}(\mathbf{e}_{t}^{\text{hard}})\right\|_2^2\right].
\end{equation}
Here, $\mathbf{sg}(\cdot)$ denotes the stop-gradient operator. This term stabilizes discrete code learning by penalizing the discrepancy between soft and hard embeddings. The second term, $\mathcal{L}_{\text{KL}}$, prevents codebook collapse by encouraging more balanced usage of the Z-token vocabulary:
\begin{equation}
\label{eq:lkl}
\mathcal{L}_{\text{KL}} =
\sum_{j=1}^{|\mathcal{V}_z|}
\hat{\mathbf{q}}_j
\Big(\log(\hat{\mathbf{q}}_j+\epsilon)-\log(u_j+\epsilon)\Big),
\end{equation}
where $\hat{\mathbf{q}}=\frac{1}{M}\sum_{m=1}^{M}\tilde{\mathbf{q}}^{(m)}$ is the average soft assignment over a minibatch of size $M$, and $u_j$ denotes the target uniform prior over $\mathcal{V}_z$. This regularizer encourages the compressor to use the extended codebook more evenly, thereby avoiding degenerate solutions in which only a small subset of Z-tokens is activated. The final training objective for the decompressor integrates the reconstruction loss with these regularization terms to ensure end-to-end consistency:
\begin{equation}
\mathcal{L}_{\text{total}} =
\mathcal{L}_{\text{tr}}
+ \lambda \mathcal{L}_{\text{KL}}
+ \beta \mathcal{L}_{\text{com}}
+ \gamma \mathcal{L}_{\text{len}},
\label{eq:total_loss}
\end{equation}
where $\lambda$, $\beta$, and $\gamma$ control the strengths of the code-usage, embedding-consistency, and length regularizers, respectively.

\subsection{Z-space inference followed by reconstruction}
\label{sec:reconstructor}
The decompressor as reconstructor paradigm positions the decompressor as a reconstructor of the input text to the compressor from Z-tokens, while a separate LLM operates in the Z-token space between compression and reconstruction to enable reasoning purely in the Z-Token space. This design decouples compression from high-level reasoning, allowing the inferencer LLM to focus on high-level semantics without handling redundant surface-level tokens. In this case, we first train the decompressor using Eqn.~\ref{eq:total_loss} with the target sequence $\mathbf{Y}$ as the compressor input $\mathbf{X}$. Then we compress both the prompt sequence $\mathbf{X}$ and the response sequence $\mathbf{Y}$ into Z-Token sequences $\mathbf{Z}^\textbf{p}$ and $\mathbf{Z}^\textbf{r}$
The separate LLM inferencer conducts autoregressive generation entirely on Z-tokens, treating them as an internal symbolic language, as: 
\begin{equation}
p_{\psi}(\mathbf{Z}^\textbf{r} \mid \mathbf{Z}^\textbf{p}) = \prod_{t=1}^{|\mathbf{Z}^\textbf{r}|} p_{\psi}(\mathbf{Z}^\textbf{r}_t \mid \mathbf{Z}^\textbf{p}, z^{\textbf{r}}_1 \dots z^{\textbf{r}}_{t-1} ),
\end{equation}
where $\psi$ represents the parameters of the inference LLM trained or adapted for symbolic reasoning. This allows operations such as summarization, continuation, and question answering to be performed in the latent space without returning to surface tokens. A key advantage of reasoning in the Z-Token space lies in its semantic density and computational efficiency. Each Z-Token encodes multiple surface tokens, enabling the model to infer long-range dependencies with significantly reduced context length. Essentially, this paradigm treats Z-tokens as a symbolic interface between compression and reasoning; the model no longer needs to reconstruct before thinking, but can think directly using its own abstract language. This provides a new window into how LLMs internally represent and manipulate semantic knowledge.

\subsection{Contextual Regularity of Z-tokens}
\label{sec:ztoken_semantics}
We observe that the learned codebook $\mathcal{V}_Z$ is reused in heterogeneous inputs: the same Z-tag index can appear in different contexts but yields different representations after decompression; conversely, the same input may have multiple valid Z sequences at different sampling temperatures. Unlike traditional tokenizers, where one token corresponds to one subword, this suggests that Z-tokens function as abstract, context-dependent semantic units, rather than deterministic mappings.

\paragraph{Qualitative illustrations.}
In several cases, Z-tokens  operate at a conceptual level rather than encoding fixed surface wording.
1) \textbf{Semantic synonymy:} sentences such as \emph{``The government implemented sweeping economic reforms to address the financial crisis''} and \emph{``Officials enacted major fiscal restructuring to mitigate the debt emergency''} share overlapping Z-tokens (e.g., $z_{42}, z_{87}$), suggesting a shared encoding of government-led economic intervention despite lexical variation in the decompressed outputs.
2) \textbf{Syntactic alternation:} \emph{``Climate change threatens agricultural productivity and requires immediate attention''} and \emph{``Agricultural productivity is threatened by climate change, which requires immediate action''} map to shared tokens (e.g., $z_{156}, z_{203}$), indicating that some codes may be relatively insensitive to surface word order while preserving higher-level semantic content.
3) \textbf{Domain-level variation:} \emph{``Advanced technology drives innovation in transportation systems''} and \emph{``Cutting-edge solutions accelerate progress in mobility infrastructure''} align on tokens such as $z_{487}, z_{99}$, again suggesting reuse at the level of abstract topical meaning rather than exact lexical realization.

\vspace{-1em}
\paragraph{Interpretation.}
These observations suggest that Z-tokens may function as reusable, underspecified semantic codes rather than fixed lexical symbols. In this view, compression abstracts away some surface variability, allowing multiple paraphrastic realizations to share parts of the same discrete representation. We stress that this is an empirical interpretation of the learned code behavior, not a claim that each Z-token carries an invariant standalone meaning.

\vspace{-1em}
\paragraph{Measuring contextual regularity.}
To assess whether this reuse is non-random, we measure the contextual consistency of each Z-token across its corpus occurrences. Let $\mathcal{O}(z_i)=\{o_1,\dots,o_{N_i}\}$ denote the set of occurrences of $z_i$ in a corpus. For each occurrence $o_p$, let $\mathbf{ctx}(o_p)$ be a fixed-width window of surface text around $o_p$, and let $\mathcal{F}(\cdot)$ be a frozen semantic encoder (e.g., BERT) that maps text to an embedding. We then define:
\begin{equation}
\label{eq:consistency}
\mathcal{C}(z_i) 
=\frac{1}{\binom{N_i}{2}} \sum_{p<q} 
\textbf{sim}\!\left( \mathcal{F}(\mathbf{ctx}(o_p)),\, \mathcal{F}(\mathbf{ctx}(o_q)) \right).
\end{equation}
A high value of $\mathcal{C}(z_i)$ indicates that occurrences of the same Z-token tend to appear in semantically similar neighborhoods, suggesting structured reuse rather than arbitrary code assignment. On our evaluation set, the mean consistency is $0.75 \pm 0.11$, which supports the claim that Z-token reuse exhibits substantial contextual regularity. This context dependence naturally raises the question of how underspecified codes are resolved during decompression. Importantly, Z-token decoding is not performed in isolation. Instead, each output token is conditioned on three sources of information: the complete Z-sequence \(Z=(z_1,\dots,z_K)\), the autoregressive text prefix \(x_{<t}\), and the base-vocabulary constraint \(\mathcal{V}_{\text{base}}\). The full Z-sequence narrows broad semantic possibilities through neighboring codes; the generated prefix progressively commits the decoder to one semantic realization; and restricting logits to the pre-trained vocabulary anchors decoding in well-formed natural language. Together, these constraints convert underspecified latent codes into context-resolved surface outputs.

\begin{figure}[!t] 
    \centering
    \includegraphics[width=0.95\textwidth]{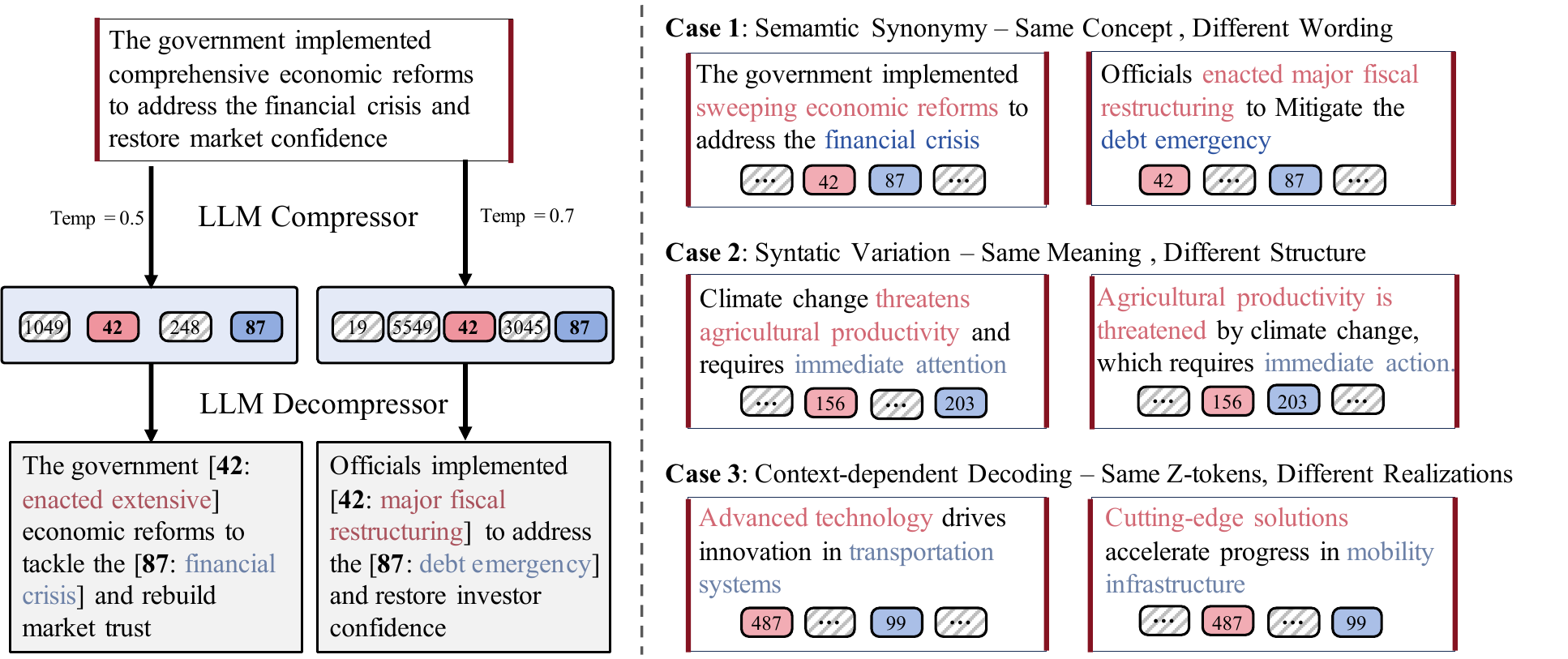}    
    \caption{The figure shows the flexibility of Z-tokens. On the \textbf{left}, the same sentence can be compressed into different Z-token sequences while preserving meaning after decompression. On the \textbf{right}, the same Z-tokens yield different outputs in different contexts. This suggests that Z-tokens are context-dependent rather than chaotic.}
    \label{fig:system_overview}
    \vspace{-2em}
\end{figure}

\vspace{-1em}
\section{Experiment}
\vspace{-1em}
\subsection{Baseline}
\vspace{-1em}
Our experiments used the Qwen3-0.6B, 1.7B and 4B. For LoRA, we apply LoRA to all projection linear layers in the LLM and set $r$ to 128, $\alpha$ to 256, and dropout to 0.1. We compared our method with state-of-the-art methods, including ICAE~\cite{incontextae}, GistToken~\cite{gisttokens}, AutoCompressor~\cite{autocompressor}, and LLOCO~\cite{lloco}: ICAE compresses the input into a fixed-length memory slot; GistToken compresses the input into a fixed-length token using attention masking; AutoCompressor compresses the input in segments into a fixed-length summary vector; and LLOCO uses a fine-tuned AutoCompressor and a base LLM for contextual compression. To ensure a fair comparison, all basic settings were kept consistent across all methods, and the Z-token size is 8k. For experimental environment information, please refer to~\ref{experiment_env}.

\vspace{-1em}
\subsection{Metrics}
\vspace{-1em}
For text reconstruction and continuation tasks, we use BLEU-4 as the core evaluation metric. Additionally, we report the perplexity score (PPL), which quantifies the difficulty of the model predicting the text. For the QuALITY dataset, we report the exact match (EM) score. For HotpotQA, NarrativeQA, and QASPER, we report the F1 score. For the CNN/DailyMail summarization task, we provide ROUGE-1, ROUGE-2, and ROUGE-L scores. 

\vspace{-1em}
\subsection{Comparison}
\vspace{-1em}
We first evaluate the model's reconstruction performance to understand whether the Z-tokens generated by the compressor contain sufficient semantic knowledge to recover the original context. We test on Wikipedia and report the BLEU-4 and CE-Loss results, as shown in Table~\ref{tab:bleu_ce}.
Our method achieves the highest BLEU-4 score and the lowest CE-Loss at various compression ratios. Notably, the actual compression ratio is even higher due to our use of autoregressive variable-length compression. For example, at a nominal compression ratio of 4$\times$, our effective average compression ratio reaches 5.1$\times$; and at a compression ratio of 8$\times$, it reaches 9.4$\times$. These results demonstrate that our variable-length compression method effectively preserves key semantic content while significantly improving compression efficiency. 


\begin{table}[t]
\centering
\noindent
\begin{minipage}[t]{0.50\linewidth}
    \vspace{0pt}
    \centering
    \includegraphics[width=\linewidth]{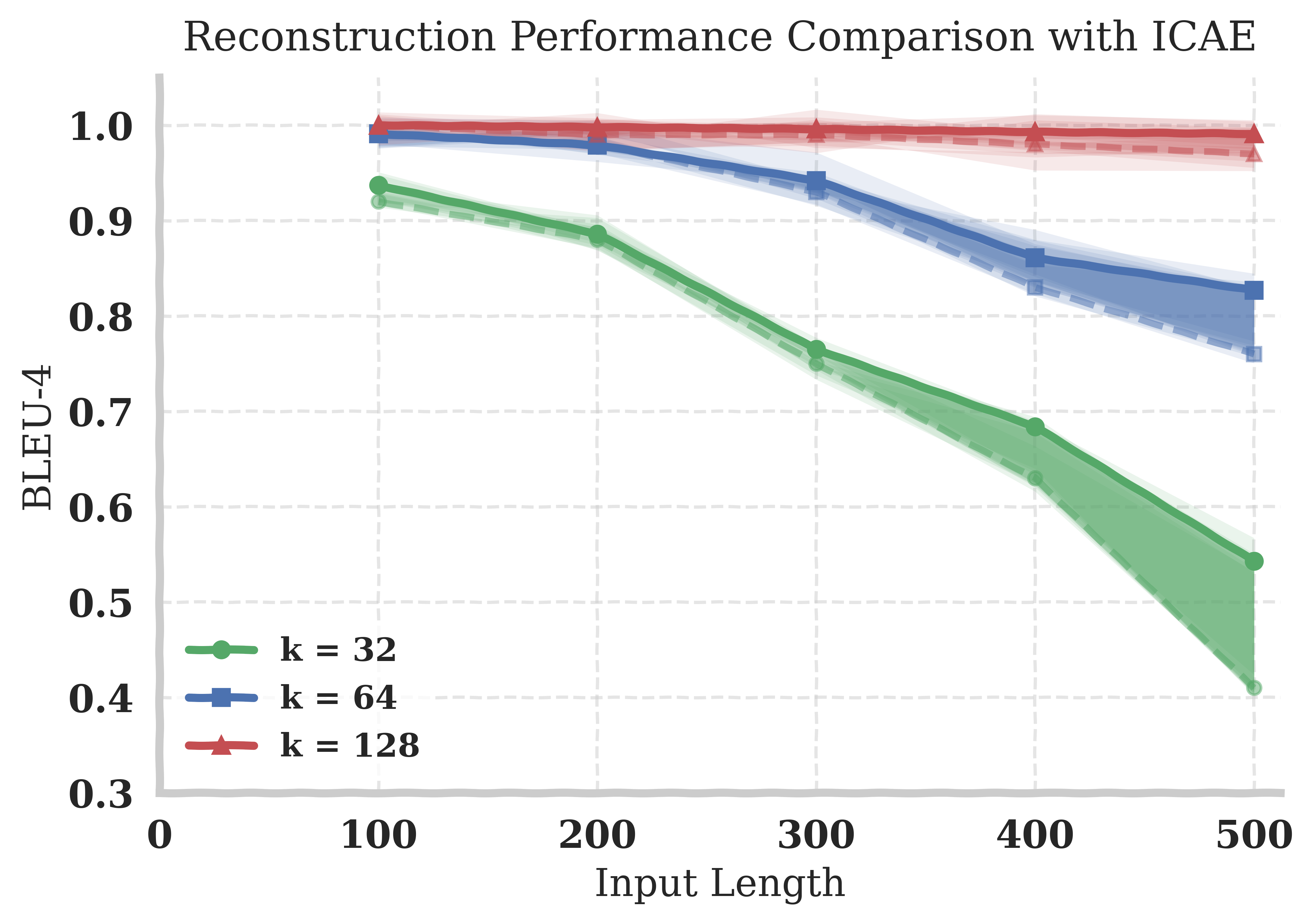}
    \captionsetup{type=figure}
    \caption{BLEU-4 at different input lengths and compression ratios. The solid line represents our method, and the dashed line represents ICAE.}
    \label{fig:placeholder}
\end{minipage}\hfill
\begin{minipage}[t]{0.47\linewidth}
    \vspace{0pt}
    \centering

    \captionsetup{type=table}
    \caption{Comparison on Wikipedia reconstruction at compression ratios of 4$\times$ and 8$\times$.}
    \label{tab:bleu_ce}
    \tiny
    \setlength{\tabcolsep}{1.8pt}
    \renewcommand{\arraystretch}{0.82}
    \resizebox{\linewidth}{!}{%
    \begin{tabular}{lcc|cc}
        \toprule
        \textbf{Method} & \multicolumn{2}{c|}{\textbf{4$\times$}} & \multicolumn{2}{c}{\textbf{8$\times$}} \\
        \cmidrule(lr){2-3}\cmidrule(lr){4-5}
        & \textbf{BLEU}$\uparrow$ & \textbf{CE}$\downarrow$ & \textbf{BLEU}$\uparrow$ & \textbf{CE}$\downarrow$ \\
        \midrule
        AutoCompressor~\cite{autocompressor}
        & 96.44 & 0.29
        & 92.67 & 0.38 \\
        Gist Token~\cite{gisttokens}
        & 96.29 & 0.21
        & 94.63 & 0.30 \\
        ICAE~\cite{incontextae}
        & 98.45 & 0.04
        & 93.44 & 0.17 \\
        \midrule
        \textbf{Ours}
        & 99.31 & 0.04
        & 96.28 & 0.13 \\
        \bottomrule
    \end{tabular}%
    }

    \vspace{0.2em}

    \captionsetup{type=table}
    \caption{Performance comparison of reconstruction task performance at different model sizes.}
    \vspace{-1em}
    \label{tab:model_scale_results}
    \tiny
    \setlength{\tabcolsep}{1.8pt}
    \renewcommand{\arraystretch}{0.90}
    \resizebox{\linewidth}{!}{%
    \begin{tabular}{lcccc}
        \toprule
        \textbf{Model} & \textbf{BLEU-4}$\uparrow$ & \textbf{PPL (Ori)}$\downarrow$ & \textbf{PPL (Z)}$\downarrow$ & \textbf{Gap} \\
        \midrule
        Qwen3-0.6B & 96.77 & 30.42 & 30.97 & +0.55 \\
        Qwen3-1.7B & 97.14 & 23.75 & 24.16 & +0.41 \\
        Qwen3-4B   & 97.57 & 16.90 & 17.14 & +0.24 \\
        \bottomrule
    \end{tabular}%
    }
\end{minipage}
\vspace{-2em}
\end{table}

\begin{wraptable}{r}{0.52\linewidth}
    \vspace{-1.1em}
    \centering
    \scriptsize
    \setlength{\tabcolsep}{3pt}
    \renewcommand{\arraystretch}{0.92}
    \caption{Comparison of PPL in the continuation task.}
    \label{tab:ppl_comparison}
    \resizebox{0.8\linewidth}{!}{%
    \begin{tabular}{lccc}
        \toprule
        \textbf{Context} & \textbf{PPL (Ori)}$\downarrow$ & \textbf{PPL (Z)}$\downarrow$ & \textbf{Gap} \\
        \midrule
        1024 $\rightarrow$ 512 & 29.96 & 30.12 & +0.16 \\
        1024 $\rightarrow$ 256 & 30.42 & 30.65 & +0.23 \\
        1024 $\rightarrow$ 128 & 29.90 & 30.34 & +0.44 \\
        \bottomrule
    \end{tabular}%
    }
    \vspace{-1.0em}
\end{wraptable}
Figure~\ref{fig:placeholder} explores the relationship between BLEU-4 and sequence length. It shows that at a compression ratio of 1/4, our method is almost unaffected, it can almost perfectly reconstruct the original input. Even at a compression ratio of 16$\times$, our method still achieves $54.3$, demonstrating the significant performance of our method. Table~\ref{tab:model_scale_results} shows the impact of different LLM sizes on the reconstruction task. It can be seen that the larger the model, the better the reconstruction performance and the lower the perplexity. The simplicity and exclusivity of the reconstruction task may lead to suboptimal generalization. Previous research~\cite{incontextae} has shown that text continuation can broadly promote the learning of more general representations because text continuation requires the model to capture not only surface-level correspondences but also the deep discourse and semantic coherence necessary for text continuation. During training, the model observes only the prefix $x_{1:t}$ of a document and learns to predict its remainder $x_{t+1:T}$. We tested this task on Wikipedia, and the results are shown in Table~\ref{tab:ppl_comparison}. The results show that our model maintains almost the same language modeling ability across different context lengths. Even when the available context is reduced from 1024 to 128, the perplexity only increases by 0.44, indicating that the model can still predict subsequent text coherently and fluently. This demonstrates that the compressed Z-token representation effectively preserves key semantics and contextual dependencies, enabling robust next-text prediction even with limited preceding context. 

The summarization task evaluates the model's ability to abstract high-level content from compressed representations. We conduct experiments on CNN/DailyMail under two paradigms, i.e., direct decompression and inference decompression. The results are shown in Table~\ref{tab:summarization_results}. We set the compression ratio to 4$\times$, however, due to variable-length compression, the average compression ratio reached 6.3$\times$, while maintaining performance comparable to or even better than the original model. Notably, in the direct decompression, the inference time was reduced from 35 min to 17 min, a speedup of 2$\times$, demonstrating the efficiency and scalability of our method.
\begin{table}[h]
\centering
\noindent
\begin{minipage}[t]{0.48\linewidth}
    \vspace{0pt}
    \centering
    \captionsetup{type=table}
    \vspace{-1em}
    \caption{Performance comparison of summarization tasks on CNN/DailyMail.}
    \label{tab:summarization_results}
    \tiny
    \setlength{\tabcolsep}{2pt}
    \renewcommand{\arraystretch}{0.90}
    \resizebox{\linewidth}{!}{%
    \begin{tabular}{lccc}
        \toprule
        \textbf{Method} & \textbf{Rouge-1}$\uparrow$ & \textbf{Rouge-2}$\uparrow$ & \textbf{Rouge-L}$\uparrow$ \\
        \midrule
        Qwen3-0.6B & 33.25 & 16.58 & 30.83 \\
        \midrule
        AutoCompressor~\cite{autocompressor} & 30.28 & 14.86 & 28.49 \\
        GistToken~\cite{gisttokens} & 29.11 & 14.57 & 27.62 \\
        ICAE~\cite{incontextae} & 30.76 & 15.31 & 28.84 \\
        \midrule
        Ours (w/o infer) & \textbf{32.58} & \textbf{16.77} & \underline{29.89} \\
        Ours (w/ infer) & \underline{32.04} & \underline{15.57} & \textbf{30.15} \\
        \bottomrule
    \end{tabular}%
    }
\end{minipage}\hfill
\begin{minipage}[t]{0.50\linewidth}
    \vspace{0pt}
    \centering
    \captionsetup{type=table}
    \vspace{-1em}
    \caption{Long-text QA results under 4$\times$ compression.}
    \label{tab:longqa}
    \tiny
    \setlength{\tabcolsep}{2.0pt}
    \renewcommand{\arraystretch}{0.90}
    \resizebox{\linewidth}{!}{%
    \begin{tabular}{lcccc}
        \toprule
        \textbf{Methods} & \textbf{QuALITY} & \textbf{HotpotQA} & \textbf{NarrativeQA} & \textbf{QASPER} \\
        \midrule
        Qwen3-0.6B~\cite{qwen3} & 38.48 & 32.18 & 16.32 & 18.44 \\
        \midrule
        AutoCom~\cite{autocompressor} & 34.43 & 23.59 & 14.41 & 16.47 \\
        ICAE~\cite{incontextae} & 34.68 & 31.65 & 15.76 & 17.62 \\
        LLOCO~\cite{lloco} & 37.33 & \textbf{33.46} & 15.88 & \underline{18.20} \\
        \midrule
        Ours (w/o infer) & \textbf{39.25} & \underline{33.35} & \textbf{16.24} & \textbf{18.31} \\
        Ours (w/ infer) & \underline{39.04} & 33.04 & \underline{15.93} & 17.66 \\
        \bottomrule
    \end{tabular}%
    }
\end{minipage}
\vspace{-1em}
\end{table}
For long text question answering, we evaluated our method on the HotpotQA, QuALITY, NarrativeQA, and QASPER datasets. We applied a 4$\times$ compression ratio to the input and tested it against two paradigms. The results are shown in Table~\ref{tab:longqa}. Under the same compression budget, our method outperforms prior compression baselines on QuALITY, NarrativeQA, and QASPER, and remains close to the best baseline on HotpotQA (33.35 vs. 33.46). We also observed that we can achieve a compression ratio of 18$\times$ for some inputs. This demonstrates that our compressed representation can preserve task-relevant semantics and maintain multi-hop inference capabilities even with significantly reduced text length.

\vspace{-1em}
\paragraph{Sliding window}

\begin{wraptable}{r}{0.28\linewidth}
    \vspace{-1.1em}
    \centering
    \small
    \setlength{\tabcolsep}{4pt}
    \renewcommand{\arraystretch}{0.98}
    \caption{QASPER sliding-window results.}
    \label{tab:qasper_f1}
    \resizebox{0.6\linewidth}{!}{%
    \begin{tabular}{lc}
        \toprule
        \textbf{Method} & \textbf{F1}$\uparrow$ \\
        \midrule
        Qwen3-0.6B & 18.44 \\
        ICAE & 17.26 \\
        \midrule
        \textbf{Ours} & \textbf{18.25} \\
        \bottomrule
    \end{tabular}%
    }
    \vspace{-1em}
\end{wraptable}
We experimented on the QASPER dataset. Due to hardware limitations, we set the window size to 1024 and the stride to 256, enabling the model to process documents incrementally while maintaining semantic continuity. For comparison, we also evaluated a baseline configuration where the entire document was fed into the model at once, as a performance cap for the uncompressed case. The results in Table~\ref{tab:qasper_f1} show that our method uses fewer tokens per inference but achieves comparable performance. This demonstrates that our compression mechanism can be naturally combined with a sliding window strategy to handle inputs exceeding the maximum context window of an LLM.

\vspace{-1em}
\subsection{Ablation Study.}

\begin{wraptable}{r}{0.45\linewidth}
    \vspace{-1.0em}
    \centering

    \captionsetup{type=table}
    \scriptsize
    \setlength{\tabcolsep}{3pt}
    \renewcommand{\arraystretch}{0.95}
    \caption{Free-decoding BLEU-4 on Wikipedia under different input lengths.}
    \label{tab:free_decode_bleu}
    \resizebox{\linewidth}{!}{%
    \begin{tabular}{lccccc}
        \toprule
        \textbf{Method} & 100 & 200 & 300 & 400 & 500 \\
        \midrule
        ICAE~\cite{incontextae} & 29.46 & 21.88 & 14.53 & 11.27 & 8.96 \\
        \midrule
        \textbf{Ours} & \textbf{58.24} & \textbf{52.23} & \textbf{46.66} & \textbf{41.84} & \textbf{37.66} \\
        \bottomrule
    \end{tabular}%
    }

    \vspace{0.8em}

    \captionsetup{type=table}
    \caption{Effect of different Z-token counts on HotpotQA.}
    \label{tab:zsize}
    \scriptsize
    \setlength{\tabcolsep}{3pt}
    \renewcommand{\arraystretch}{0.90}
    \resizebox{0.95\linewidth}{!}{%
    \begin{tabular}{l|ccccc}
        \toprule
        \textbf{Dataset} & \textbf{1k} & \textbf{2k} & \textbf{4k} & \textbf{8k} & \textbf{10k} \\
        \midrule
        HotpotQA (F1) & 16.23 & 22.45 & 27.96 & \textbf{33.35} & 33.27 \\
        \bottomrule
    \end{tabular}%
    }

    \vspace{-1em}
\end{wraptable}
We first study the effect of scheduled sampling on free-decoding performance in Wikipedia reconstruction. We compare against ICAE, which is trained with teacher forcing and therefore suffers from exposure bias: although decoding is accurate under teacher-forced training, errors accumulate rapidly during free decoding. To improve robustness, we adopt scheduled sampling in training (Section~\ref{sec:decompressor}), gradually increasing the sampling probability from 0 to 0.5. We do not push $p$ to 1, since more aggressive sampling can further stabilize free decoding but often slows convergence and degrades fluency. Under this setting, Table~\ref{tab:free_decode_bleu} shows that scheduled sampling substantially improves free-decoding reconstruction, making the model more robust to autoregressive error accumulation. We further study the effect of Z-token vocabulary size on HotpotQA using settings of 1k, 2k, 4k, 8k, and 10k, as reported in Table~\ref{tab:zsize}. When the Z-token set is too small, the compressed representation becomes overly restrictive and fails to preserve enough relational and contextual information, leading to worse QA performance. As the vocabulary size increases, performance improves accordingly, but the gain becomes marginal beyond 8k, suggesting that the representation has already captured the key reasoning information. This result indicates a practical trade-off between compression capacity and computational cost. Given the highly structured nature of Wikipedia data, we also tested it on the less structured HumanEval; the results are shown in Appendix~\ref{humaneval}, demonstrating the robustness of our method.

\vspace{-1em}
\paragraph{Practical Efficiency Analysis}

\begin{table}[t]
\centering
\noindent
\begin{minipage}[t]{0.49\linewidth}
    \vspace{0pt}
    \centering

    \captionsetup{type=table}
    \caption{Results under different KV-cache budgets / compression ratios. A: Acc, M: memory (GB), T: inference time (s).}
    \label{tab:h2o_budget_compact_sharedbase_perdataset}
    \tiny
    \setlength{\tabcolsep}{1.4pt}
    \renewcommand{\arraystretch}{0.90}
    \resizebox{0.98\linewidth}{!}{%
    \begin{tabular}{l l c c c}
        \toprule
        \textbf{Data} & \textbf{Method} & \textbf{25\%} & \textbf{50\%} & \textbf{Base} \\
        \cmidrule(lr){3-3}\cmidrule(lr){4-4}\cmidrule(lr){5-5}
         & & \multicolumn{1}{c}{\textbf{A/M/T}} &
             \multicolumn{1}{c}{\textbf{A/M/T}} &
             \multicolumn{1}{c}{\textbf{A/M/T}} \\
        \midrule
        \multirow{2}{*}{PIQA}
        & Ours & 62.70/10.63/142.57 & 66.53/14.28/174.32 & \multirow{2}{*}{67.30/27.74/198.26} \\
        & H2O  & 61.48/11.49/134.28 & 65.32/16.91/170.21 & \\
        \midrule
        \multirow{2}{*}{RTE}
        & Ours & 68.46/9.30/56.88 & 71.70/14.20/70.28 & \multirow{2}{*}{73.64/25.41/82.12} \\
        & H2O  & 68.01/10.83/58.46 & 71.35/15.16/66.88 & \\
        \bottomrule
    \end{tabular}%
    }


\end{minipage}\hfill
\begin{minipage}[t]{0.49\linewidth}
    \vspace{0pt}
    \centering

    \captionsetup{type=table}
    \caption{Attention/MLP time ratio at different input lengths for Qwen3-4B and 8B. Higher attention ratios at longer inputs indicate that reducing effective context length remains beneficial even for larger models.}
    \label{tab:attn_mlp_time_ratio}
    \small
    \setlength{\tabcolsep}{4pt}
    \renewcommand{\arraystretch}{0.95}
    \resizebox{0.98\linewidth}{!}{%
    \begin{tabular}{l c c}
        \toprule
        \textbf{Model} & \textbf{Input} & \textbf{Attn / MLP (\%)} \\
        \midrule
        Qwen3-8B & $8\times256/512/1024$ & 43.8/47.9/54.4 \\
        Qwen3-4B & $8\times256/512/1024$ & 47.7/52.9/60.4 \\
        \bottomrule
    \end{tabular}%
    }
\end{minipage}
\vspace{-1.5em}
\end{table}

A natural question is whether context compression remains useful as model size increases. In larger models, a greater fraction of runtime may shift from attention to MLP computation, which could potentially weaken the benefit of shortening the input context. To examine this issue, we report the Attention/MLP time ratio for Qwen3-4B and Qwen3-8B at different input lengths in Table~\ref{tab:attn_mlp_time_ratio}. The results show that, although MLP computation becomes increasingly important in larger models, attention still occupies a substantial fraction of runtime and grows consistently with sequence length. In particular, as the input length increases from $8\times256$ to $8\times1024$, the attention ratio rises markedly for both Qwen3-4B and Qwen3-8B. This indicates that shortening the effective context remains beneficial even at larger model scales, and therefore the proposed compression mechanism retains its practical value beyond small or medium-sized LLMs.

\vspace{-1em}
\paragraph{Inference efficiency analysis}
\begin{wraptable}{r}{0.52\linewidth}
    \vspace{-1.0em}
    \centering
    \scriptsize
    \caption{On Qwen3-4B, we compare long-context generation with and without compression, reporting inference time, throughput, and peak GPU memory.}
    \label{tab:cached_efficiency}
    \setlength{\tabcolsep}{3pt}
    \renewcommand{\arraystretch}{0.95}
    \resizebox{\linewidth}{!}{%
    \begin{tabular}{c|ccc|ccc}
    \toprule
    \multirow{2}{*}{\textbf{Input Length}} & \multicolumn{3}{c|}{\textbf{Baseline}} & \multicolumn{3}{c}{\textbf{Ours (Compressed Context)}} \\
    \cmidrule(lr){2-4}\cmidrule(lr){5-7}
     & \textbf{Time (s)} & \textbf{Tok/s} & \textbf{Mem. (GB)} & \textbf{Time (s)} & \textbf{Tok/s} & \textbf{Mem. (GB)} \\
    \midrule
    10k  & 11.36 & 45.09 & 24.86 & 9.25 & 70.61 & 12.77 \\
    20k  & 16.11 & 34.06 & 41.82 & 12.39 & 56.28 & 19.17 \\
    30k & 22.67 & 24.09 & 56.48 & 17.21 & 47.69 & 25.40 \\
    \bottomrule
    \end{tabular}%
    }
    \vspace{-1.0em}
\end{wraptable}

To evaluate the practical efficiency of compressed representations, we compare standard long-context generation with generation from compressed Z-token sequences on Qwen3-4B under the same output length of 64 tokens. For our method, the reported time includes both the autoregressive compression step and the decompression/generation step, while throughput and memory are measured during the generation stage. As shown in Table~\ref{tab:cached_efficiency}, our method consistently reduces end-to-end latency and generation-stage peak GPU memory across all input lengths, while improving generation throughput. The advantage becomes more pronounced as the context grows. At 30k tokens, latency decreases from 22.67\,s to 17.21\,s, throughput increases from 24.09 to 47.69 tokens/s, and peak memory decreases from 56.48\,GB to 25.40\,GB. These results suggest that Z-token compression remains beneficial even when compression overhead is included, especially for long-context and memory-constrained inference.

\vspace{-1em}
\paragraph{Adaptive Z-token allocation.}

\begin{wraptable}{r}{0.48\linewidth}
    \vspace{-1.0em}
    \centering
    \scriptsize
    \caption{Correlation between allocated Z-token count and semantic-density proxies.}
    \label{tab:semantic_density_corr_main}
    \setlength{\tabcolsep}{3pt}
    \renewcommand{\arraystretch}{0.90}
    \resizebox{\linewidth}{!}{%
    \begin{tabular}{llcc}
    \toprule
    \textbf{Dataset} & \textbf{Proxy} & \textbf{Pearson $r$} & \textbf{Spearman $\rho$} \\
    \midrule
    Wikipedia & Surprisal           & 0.54 & 0.57 \\
    HotpotQA  & Surprisal           & 0.49 & 0.52 \\
    Wikipedia & Entity-rare density & 0.36 & 0.39 \\
    HotpotQA  & Entity-rare density & 0.43 & 0.46 \\
    \bottomrule
    \end{tabular}%
    }
    \vspace{-2.0em}
\end{wraptable}
To examine whether variable-length compression reflects information density rather than nearly uniform allocation, we conduct a controlled analysis under the same nominal $4\times$ compression setting as the main experiments. Using 10,000 fixed-length segments from Wikipedia and HotpotQA, we analyze the relationship between allocated Z-token count and two segment-level proxies of semantic density: surprisal and entity-rare density. Since all segments have the same length, the observed trends cannot be attributed to trivial length effects.
As shown in Table~\ref{tab:semantic_density_corr_main}, the allocated Z-token count is positively correlated with both proxies across both domains. Moreover, the trend across surprisal buckets is monotonic: segments with higher surprisal consistently receive more latent tokens. These results suggest that the learned compressor does not allocate latent budget uniformly, but instead assigns more Z-tokens to segments that are harder to predict or contain denser factual content. This provides empirical support for the intended allocation principle behind our variable-length design. Full details and additional results are given in Appendix~\ref{app:semantic_density}.

\vspace{-1em}
\subsection{Conclusion.} 
\vspace{-1em}
This paper proposes a discretization method that integrates compression and generation into one LLM, enabling the model to naturally compress long texts into variable-length, controllable Z-tokens before downstream tasks. Extensive experiments on multiple tasks demonstrate that the method significantly reduces the latency and memory cost of downstream long-context inference while maintaining good task performance. These results show that allowing the LLM to infer its own compact latent language provides a promising direction for efficient long-context modeling. Future work could consider developing a global coordination mechanism across windows and a task-aware compression strategy to better preserve the structure required for specific downstream objectives.

\newpage
{
\bibliographystyle{unsrt}
\bibliography{neurips_2026}
}


\newpage
\appendix

\section{Implementation details.}
\label{experiment_env}
All experiments used six NVIDIA GPUs: two A800-80G and four V100-32G. Each configuration is equipped with one Intel Xeon Gold 6348 CPU. The software environment includes Ubuntu 22.04 LTS, Python 3.12, PyTorch 2.5.1, and CUDA 12.4. Unless otherwise specified, the base LLM used in the experiments is Qwen3-0.6B.

During the reconstruction and continuation of pre-training, we used Qwen3-0.6B as the base model and trained the compressor-decompressor framework using LoRA. Training was performed on two NVIDIA A800 GPUs with a precision of bfloat16, and FlashAttention and gradient checkpointing were enabled to improve memory efficiency. The batch size per GPU was set to 4, resulting in a global batch size of 8. The input length of the pre-training snippets was fixed at 1024 tokens. The overall training time was approximately 3-4 days. While this training scale is significantly larger than standard downstream fine-tuning, it is appropriate for our setup because the model needs to adapt not only to supervised tasks but also to achieve stable compression-decompression behavior on discrete, variable-length latent sequences.




\section{Generalization}
\label{humaneval}
\begin{wraptable}{r}{0.42\textwidth}
    \vspace{-0.8em}
    \centering
    \scriptsize
    \caption{HumanEval results, reported with pass@1.}
    \label{tab:humaneval_main}
    \setlength{\tabcolsep}{3pt}
    \renewcommand{\arraystretch}{0.90}
    \begin{tabular}{lccc}
    \toprule
    \textbf{Dataset} & \textbf{Base Model} & \textbf{AutoCompressor} & \textbf{Ours} \\
    \midrule
    HumanEval & 18.90 & 12.64 & 16.85 \\
    \bottomrule
    \end{tabular}
    \vspace{-1.0em}
\end{wraptable}

We further evaluate the generalization ability of our method on HumanEval, which contains relatively unstructured code-generation prompts and is therefore substantially different from the long-form text corpora used in pretraining. As shown in Table~\ref{tab:humaneval_main}, under $4\times$ compression, AutoCompressor leads to a noticeable drop in pass@1 accuracy, whereas our method preserves performance more effectively and remains substantially closer to the base model. These results suggest that the learned compression mechanism generalizes beyond standard document reconstruction and can retain useful information even in less structured input domains.

\section{Analyzing Variable-Length Z-Token Allocation}
\label{app:semantic_density}

A central question behind variable-length compression is whether the compressor allocates more Z-tokens to semantically denser segments, rather than producing lengths that are nearly uniform or driven by superficial statistics. To probe this behavior, we conduct a controlled diagnostic study calibrated to the main experimental regime, namely nominal $4\times$ compression. We emphasize that this analysis is intended as a targeted diagnostic of allocation behavior under the same compression regime as the main experiments, rather than as a definitive annotation-based measure of semantic density.

We construct 10{,}000 diagnostic segments, including 5000 Wikipedia-style segments from reconstruction scenarios and 5000 HotpotQA-style segments from long-context question answering scenarios. Each segment contains exactly 256 input tokens. For each segment $i$, we record the allocated Z-token count $k_i$ together with two segment-level proxies of semantic density:
\begin{itemize}
    \item \textbf{Segment surprisal}: the mean token-level negative log-likelihood under the base model;
    \item \textbf{Entity-rare density}: the fraction of named entities and rare words in the segment.
\end{itemize}
We then report Pearson and Spearman correlations between each proxy and $k_i$, and additionally summarize the mean allocated Z-token count across surprisal quintiles. Because all segments have the same input length, the observed trends cannot be explained by trivial dependence on raw segment length. We adopt this controlled protocol to isolate allocation behavior under matched segment length and compression budget, which would otherwise be more difficult to disentangle from confounding factors in full end-to-end task traces.

\begin{table}[h]
\centering
\scriptsize
\caption{Correlation between allocated Z-token count and semantic-density proxies in the controlled diagnostic study.}
\label{tab:semantic_density_corr}
\setlength{\tabcolsep}{3pt}
\renewcommand{\arraystretch}{0.90}
\begin{tabular}{llcc}
\toprule
\textbf{Dataset} & \textbf{Proxy} & \textbf{Pearson $r$} & \textbf{Spearman $\rho$} \\
\midrule
Wikipedia & Segment surprisal   & 0.54 & 0.57 \\
HotpotQA  & Segment surprisal   & 0.49 & 0.52 \\
Wikipedia & Entity-rare density & 0.36 & 0.39 \\
HotpotQA  & Entity-rare density & 0.43 & 0.46 \\
\bottomrule
\end{tabular}
\vspace{-0.8em}
\end{table}

\begin{figure}[h]
\centering
\scriptsize
\setlength{\tabcolsep}{4pt}
\renewcommand{\arraystretch}{0.92}
\begin{tabular}{lcc}
\toprule
\textbf{Surprisal quintile} & \textbf{Mean allocated Z-tokens} & \textbf{Trend bar} \\
\midrule
Q1 (lowest)  & 37.1 & \rule{0.93cm}{1.0ex} \\
Q2           & 41.0 & \rule{1.03cm}{1.0ex} \\
Q3           & 46.4 & \rule{1.16cm}{1.0ex} \\
Q4           & 52.2 & \rule{1.31cm}{1.0ex} \\
Q5 (highest) & 59.8 & \rule{1.50cm}{1.0ex} \\
\bottomrule
\end{tabular}
\caption{Binned trend plot showing the relationship between surprisal quintile and allocated Z-token count. Higher-surprisal segments consistently receive more Z-tokens.}
\label{fig:semantic_density_trend}
\vspace{-1.0em}
\end{figure}

Across both domains, the allocated Z-token count is positively associated with multiple semantic-density proxies, and the surprisal-bucket trend is monotonic. These results provide supporting evidence that Z-token allocation is correlated with segment-level information density, rather than being approximately uniform across segments. At the same time, we do not claim that these proxies fully capture semantic complexity, nor that this controlled analysis replaces a fully annotated semantic-density evaluation. Instead, the results should be understood as evidence that the learned variable-length compressor behaves consistently with the intended allocation principle: segments that are harder to predict or contain denser factual content tend to receive a larger share of the latent Z-token budget.

\section{Variable-length compression.}
\label{thero}
Although our compressor is implemented autoregressively, its behavior can be understood from a simple constrained information allocation perspective. The key question is not merely how to compress an input sequence $X=\{x_1,\dots,x_N\}$ into a shorter latent sequence $Z=\{z_1,\dots,z_K\}$, but how to allocate a limited compression budget across parts of the input so that semantic distortion is minimized.

Consider partitioning $X$ into $M$ local segments,
\[
X = \{X^{(1)}, X^{(2)}, \dots, X^{(M)}\},
\]
and suppose the compressor allocates $K_i$ latent tokens to segment $X^{(i)}$, with
\[
\sum_{i=1}^M K_i = K.
\]
Let $D_i(K_i)$ denote the reconstruction or task-induced distortion contributed by segment $X^{(i)}$ when it is represented by $K_i$ Z-tokens. A natural objective is then
\[
\min_{\{K_i\}} \sum_{i=1}^M D_i(K_i)
\quad
\text{s.t.}
\quad
\sum_{i=1}^M K_i \le \frac{N}{r},
\]
where $r$ is the target compression ratio. This is a standard budgeted allocation problem: the model must decide where the limited latent capacity should be spent.

To make this objective interpretable, we introduce a segment-wise semantic complexity measure
\[
H_i \;\triangleq\; H\!\left(X^{(i)} \mid X^{(<i)}\right),
\]
which can be viewed as the conditional uncertainty or information density of segment $X^{(i)}$ given its preceding context. Intuitively, segments with larger $H_i$ are harder to predict from context and therefore require more latent capacity to preserve their semantics. We assume that the distortion for each segment decreases with the number of allocated Z-tokens and that the reduction is larger for semantically denser segments. A simple form satisfying this intuition is
\[
D_i(K_i) = H_i \exp(-\beta K_i),
\]
where $\beta > 0$ controls how efficiently each additional Z-token reduces distortion.

Under this model, the constrained problem becomes
\[
\min_{\{K_i\}} \sum_{i=1}^M H_i \exp(-\beta K_i)
\quad
\text{s.t.}
\quad
\sum_{i=1}^M K_i \le \frac{N}{r}.
\]
Its Lagrangian is
\[
\mathcal{L}(\{K_i\},\lambda)
=
\sum_{i=1}^M H_i \exp(-\beta K_i)
+
\lambda \left(\sum_{i=1}^M K_i - \frac{N}{r}\right),
\]
with $\lambda \ge 0$. Taking derivatives with respect to $K_i$ gives
\[
\frac{\partial \mathcal{L}}{\partial K_i}
=
-\beta H_i \exp(-\beta K_i) + \lambda.
\]
At the optimum, for every active segment,
\[
\beta H_i \exp(-\beta K_i^\star) = \lambda,
\]
which yields
\[
K_i^\star
=
\frac{1}{\beta}
\left(
\log(\beta H_i) - \log \lambda
\right).
\]
Therefore,
\[
K_i^\star \text{ is monotonically increasing in } H_i.
\]
This directly implies that, under a fixed global compression budget, semantically denser segments should receive more Z-tokens, while more predictable or redundant segments should receive fewer.

This view provides a theoretical interpretation of our design. The autoregressive compressor with the special token $[\mathrm{EOS}\text{-}Z]$ determines the realized length $K$ adaptively, while the length regularizer
\[
L_{\text{len}}=\left(\frac{K}{N}-\frac{1}{r}\right)^2
\]
acts as a soft relaxation of the hard budget constraint $\sum_i K_i \le N/r$. In other words, our training objective can be viewed as optimizing a soft rate--distortion trade-off: the model is encouraged to minimize downstream reconstruction or task loss while spending its latent Z-token budget where it reduces semantic distortion the most.

Importantly, this analysis does not claim that each Z-token corresponds to a fixed symbolic meaning, nor that the trained transformer exactly solves the above optimization. Rather, it explains why variable-length compression is the natural outcome of an adaptive latent bottleneck: if different parts of the input carry different amounts of conditional information, then an optimal compressor should not allocate the same number of latent tokens everywhere. Our method implements this principle in a discrete autoregressive form that is compatible with the native generation process of LLMs.

\section{Dataset statistics.}
Table ~\ref{tab:dataset_stats} summarizes the main downstream long-context datasets used in our experiments. These benchmarks cover a variety of tasks, including abstract text summarization (CNN/DailyMail), multi-hop question answering (HotpotQA), long narrative comprehension (NarrativeQA), scientific document question answering (QASPER), and the more challenging long-context multiple-choice reading comprehension (QUALITY). The reported dataset sizes follow standard publicly available partitions.

\begin{table*}[h]
\centering
\caption{Statistics of the downstream long-context datasets used in our experiments. The reported sizes follow standard public splits; exact numbers may vary slightly depending on preprocessing.}
\label{tab:dataset_stats}
\setlength{\tabcolsep}{6pt}
\renewcommand{\arraystretch}{1.08}
\begin{tabular*}{\textwidth}{@{\extracolsep{\fill}}lcccccc}
\hline
\textbf{Dataset} & \textbf{Task} & \textbf{Train} & \textbf{Valid} & \textbf{Test} & \textbf{Output} \\
\hline
CNN/DailyMail & Summarization     & 287{,}227 & 13{,}368 & 11{,}490 & Summary \\
HotpotQA      & Multi-hop QA      & 90{,}447  & 7{,}405  & 7{,}405  & Answer \\
NarrativeQA   & Long-form QA      & 32{,}747  & 3{,}461  & 10{,}557 & Answer \\
QASPER        & Scientific QA     & 2{,}593   & 1{,}006  & 1{,}451  & Answer \\
QuALITY       & Multiple-choice QA& 265       & 265      & 265      & Option \\
\hline
\end{tabular*}
\end{table*}

During the reconstruction and continuation pre-training phases, we used a large subset of Wikipedia containing approximately 300,000 full-text articles randomly selected from various domains. To construct long-context training instances, each article was divided into blocks of 1024 units in length, with 256 overlapping terms between adjacent blocks within the same document to maintain local semantic continuity. In practice, each article generated an average of about 3 to 6 training segments, totaling approximately 1.45 million segments.

\section{Training and Inference Pipeline}

\begin{algorithm}[H]
\caption{Training the Z-token compressor, decompressor, and inferencer}
\label{alg:training}
\begin{algorithmic}[1]
\Require Pretrained LLM backbone $f_{\Theta}$; base vocabulary $V_{\mathrm{base}}$; Z-token vocabulary $V_Z$; training examples $(X,Y)$; target compression ratio $r$.
\State Initialize three LoRA adapter sets: compressor adapter $\Delta_{\phi}$, decompressor adapter $\Delta_{\theta}$, and inferencer adapter $\Delta_{\psi}$.
\State Extend the output vocabulary with $V_Z$ for compressor and inferencer generation.
\For{each minibatch}
    \State \textbf{Compression:} activate $\Delta_{\phi}$ and generate a variable-length Z-token sequence
    \[
        Z = (z_1,\ldots,z_K), \quad z_t \in V_Z,
    \]
    by autoregressive decoding from $p_{\phi}(Z \mid X)$ until \texttt{[EOS-Z]} is emitted.
    \State Use Gumbel-Softmax with a straight-through estimator to obtain differentiable Z-token embeddings during training.
    \State Compute the length regularization term
    \[
        \mathcal{L}_{\mathrm{len}} =
        \left(\frac{K}{|X|} - \frac{1}{r}\right)^2 .
    \]
    \State \textbf{Decompression / reconstruction:} activate $\Delta_{\theta}$ and decode from $Z$ to the target natural-language sequence.
    \If{reconstruction or continuation training}
        \State Set the target sequence to the original or continued text and compute
        \[
            \mathcal{L}_{\mathrm{tr}}
            =
            - \sum_t \log p_{\theta}(y_t \mid y_{<t}, Z).
        \]
    \ElsIf{downstream supervised training}
        \State Set the target sequence to the task output $Y$ and compute the same conditional generation loss.
    \EndIf
    \State Add codebook usage and commitment regularizers $\mathcal{L}_{\mathrm{KL}}$ and $\mathcal{L}_{\mathrm{com}}$.
    \State Update $\Delta_{\phi}$ and $\Delta_{\theta}$ using
    \[
        \mathcal{L}_{\mathrm{comp-dec}}
        =
        \mathcal{L}_{\mathrm{tr}}
        + \lambda \mathcal{L}_{\mathrm{KL}}
        + \beta \mathcal{L}_{\mathrm{com}}
        + \gamma \mathcal{L}_{\mathrm{len}} .
    \]
    \State \textbf{Optional Z-space inference training:} compress both the prompt/input $X$ and the target response $Y$ into $Z^p$ and $Z^r$ using $\Delta_{\phi}$.
    \State Activate $\Delta_{\psi}$ and train the inferencer to predict the response codes:
    \[
        \mathcal{L}_{\mathrm{infer}}
        =
        - \sum_t \log p_{\psi}(z^r_t \mid Z^p, z^r_{<t}) .
    \]
\EndFor
\end{algorithmic}
\end{algorithm}

\begin{algorithm}[H]
\caption{Inference with adapter switching}
\label{alg:inference}
\begin{algorithmic}[1]
\Require Input context or prompt $X$; pretrained LLM backbone $f_{\Theta}$; LoRA adapters $\Delta_{\phi}$, $\Delta_{\theta}$, $\Delta_{\psi}$.
\State \textbf{Step 1: Compress.} Activate compressor adapter $\Delta_{\phi}$ and generate
\[
    Z^p \sim p_{\phi}(Z \mid X), \quad Z^p \subset V_Z .
\]
\State \textbf{Mode A: Direct decompression / direct task inference.}
\State Activate decompressor adapter $\Delta_{\theta}$ and generate the output in the base vocabulary:
\[
    \hat{Y} \sim p_{\theta}(Y \mid Z^p), \quad \hat{Y} \subset V_{\mathrm{base}} .
\]
\State \textbf{Mode B: Z-space autoregressive inference.}
\State Activate inferencer adapter $\Delta_{\psi}$ and generate response Z-tokens:
\[
    \hat{Z}^r \sim p_{\psi}(Z^r \mid Z^p), \quad \hat{Z}^r \subset V_Z .
\]
\State Activate decompressor adapter $\Delta_{\theta}$ and decode the response codes into natural language:
\[
    \hat{Y} \sim p_{\theta}(Y \mid \hat{Z}^r), \quad \hat{Y} \subset V_{\mathrm{base}} .
\]
\State \Return $\hat{Y}$.
\end{algorithmic}
\end{algorithm}

All three functional roles are implemented by the same pretrained LLM backbone with different LoRA adapters. The compressor, decompressor, and Z-space inferencer have their own adapter parameters, denoted by $\Delta_{\phi}$, $\Delta_{\theta}$, and $\Delta_{\psi}$, respectively. At inference time, the backbone parameters are shared and frozen, and the model switches between these LoRA adapters depending on the current stage. This design avoids maintaining three independent full LLMs while still allowing each role to specialize in a different mapping: natural language to Z-tokens, Z-tokens to natural language, and Z-tokens to Z-tokens.

\newpage

\section*{NeurIPS Paper Checklist}

\begin{enumerate}

\item {\bf Claims}
    \item[] Question: Do the main claims made in the abstract and introduction accurately reflect the paper's contributions and scope?
    \item[] Answer: \answerYes{}{} 
    \item[] Justification: The abstract and introduction clearly present the main contribution of the paper: adapting an off-the-shelf LLM into a token compressor and decompressor via variable-length discrete latent Z-tokens, and the empirical sections support these claims across reconstruction and downstream tasks.

\item {\bf Limitations}
    \item[] Question: Does the paper discuss the limitations of the work performed by the authors?
    \item[] Answer: \answerYes{}{} 
    \item[] Justification: The paper discusses practical scope and remaining challenges, including hardware constraints, sliding-window settings, model scale effects, and future directions such as global coordination across windows and task-aware compression, which together clarify current limitations and scope.

\item {\bf Theory assumptions and proofs}
    \item[] Question: For each theoretical result, does the paper provide the full set of assumptions and a complete (and correct) proof?
    \item[] Answer: \answerYes{}
    \item[] Justification: The paper states the assumptions underlying the theoretical motivation for variable-length compression in the main text and provides further derivation and discussion in the appendix.

    \item {\bf Experimental result reproducibility}
    \item[] Question: Does the paper fully disclose all the information needed to reproduce the main experimental results of the paper to the extent that it affects the main claims and/or conclusions of the paper (regardless of whether the code and data are provided or not)?
    \item[] Answer: \answerYes{}
    \item[] Justification: The paper describes the model setup, training framework, datasets, baselines, compression settings, LoRA configuration, evaluation metrics, and implementation environment in sufficient detail to understand and reproduce the main experiments.

\item {\bf Open access to data and code}
    \item[] Question: Does the paper provide open access to the data and code, with sufficient instructions to faithfully reproduce the main experimental results, as described in supplemental material?
    \item[] Answer: \answerYes{}
    \item[] Justification: We will release the code and provide sufficient instructions, including the environment setup, data preparation, and commands needed to reproduce the main experimental results.

\item {\bf Experimental setting/details}
    \item[] Question: Does the paper specify all the training and test details (e.g., data splits, hyperparameters, how they were chosen, type of optimizer) necessary to understand the results?
    \item[] Answer: \answerYes{}
    \item[] Justification: The paper reports the training and evaluation settings needed to understand the experiments, including model sizes, LoRA configuration, compression ratios, window sizes, metrics, hardware setup, and core implementation choices, with additional details provided in the appendix.

\item {\bf Experiment statistical significance}
    \item[] Question: Does the paper report error bars suitably and correctly defined or other appropriate information about the statistical significance of the experiments?
    \item[] Answer: \answerYes{}
    \item[] Justification: The paper states that experiments were independently repeated three times and that the reported scores are averaged across runs, providing information about result stability.

\item {\bf Experiments compute resources}
    \item[] Question: For each experiment, does the paper provide sufficient information on the computer resources (type of compute workers, memory, time of execution) needed to reproduce the experiments?
    \item[] Answer: \answerYes{}
    \item[] Justification: The appendix specifies the compute environment, including GPU types, CPU, operating system, software stack, training precision, and approximate training time.
    
\item {\bf Code of ethics}
    \item[] Question: Does the research conducted in the paper conform, in every respect, with the NeurIPS Code of Ethics \url{https://neurips.cc/public/EthicsGuidelines}?
    \item[] Answer: \answerYes{}
    \item[] Justification: To the best of our knowledge, the work conforms to the NeurIPS Code of Ethics.

\item {\bf Broader impacts}
    \item[] Question: Does the paper discuss both potential positive societal impacts and negative societal impacts of the work performed?
    \item[] Answer: \answerNo{}
    \item[] Justification: The paper is primarily a methodological contribution focused on efficient long-context modeling and does not include a dedicated broader impact discussion.
    
\item {\bf Safeguards}
    \item[] Question: Does the paper describe safeguards that have been put in place for responsible release of data or models that have a high risk for misuse (e.g., pre-trained language models, image generators, or scraped datasets)?
    \item[] Answer: \answerNA{}
    \item[] Justification: The paper does not release a high-risk dataset or model requiring special safeguards as part of the submission.

\item {\bf Licenses for existing assets}
    \item[] Question: Are the creators or original owners of assets (e.g., code, data, models), used in the paper, properly credited and are the license and terms of use explicitly mentioned and properly respected?
    \item[] Answer: \answerYes{}
    \item[] Justification: The paper properly cites the pretrained models, datasets, and prior methods used in experiments, and uses standard research assets in a conventional academic setting.

\item {\bf New assets}
    \item[] Question: Are new assets introduced in the paper well documented and is the documentation provided alongside the assets?
    \item[] Answer: \answerNA{}
    \item[] Justification: The paper does not present a new released dataset or benchmark asset as a primary contribution.

\item {\bf Crowdsourcing and research with human subjects}
    \item[] Question: For crowdsourcing experiments and research with human subjects, does the paper include the full text of instructions given to participants and screenshots, if applicable, as well as details about compensation (if any)? 
    \item[] Answer: \answerNA{}
    \item[] Justification: The paper does not involve crowdsourcing or research with human subjects.

\item {\bf Institutional review board (IRB) approvals or equivalent for research with human subjects}
    \item[] Question: Does the paper describe potential risks incurred by study participants, whether such risks were disclosed to the subjects, and whether Institutional Review Board (IRB) approvals (or an equivalent approval/review based on the requirements of your country or institution) were obtained?
    \item[] Answer: \answerNA{}
    \item[] Justification: The paper does not involve crowdsourcing or research with human subjects.

\item {\bf Declaration of LLM usage}
    \item[] Question: Does the paper describe the usage of LLMs if it is an important, original, or non-standard component of the core methods in this research? Note that if the LLM is used only for writing, editing, or formatting purposes and does \emph{not} impact the core methodology, scientific rigor, or originality of the research, declaration is not required.
    \item[] Answer: \answerYes{}
    \item[] Justification: The paper explicitly describes the use of pretrained LLMs as the core methodological component, including their roles as compressor, decompressor, and inferencer modules.

\end{enumerate}

\end{document}